\documentclass[unnumsec,webpdf,contemporary,large,namedate]{custom}%

\graphicspath{{Fig/}}

\theoremstyle{thmstyleone}%
\theoremstyle{thmstyletwo}%
\theoremstyle{thmstylethree}%

\usepackage{graphicx}
\usepackage{color}
\usepackage{xcolor}
\usepackage{booktabs}
\usepackage{multirow}
\usepackage{hyperref}
\usepackage{caption}
\usepackage{amsmath}
\usepackage{amsfonts, amssymb}
\usepackage{colortbl}
\usepackage{subfigure}

\definecolor{purple}{RGB}{128, 0, 128}
\definecolor{LightRed}{rgb}{1,0.92,0.92}
\definecolor{LightOrange}{rgb}{1,0.95,0.88}
\definecolor{LightYellow}{rgb}{1.0,1.0,0.84}
\definecolor{LightGreen}{rgb}{0.9,1.0,0.88}
\definecolor{LightCyan}{rgb}{0.9,1,1}
\definecolor{LightBlue}{rgb}{0.9,0.94,1}
\definecolor{LightIndigo}{rgb}{0.92,0.9,1}
\definecolor{LightMagenta}{rgb}{0.96,0.86,1}
\definecolor{DirtyWhite}{rgb}{0.96,0.96,0.96}

\DeclareSymbolFont{extraup}{U}{zavm}{m}{n}
\DeclareMathSymbol{\varheart}{\mathalpha}{extraup}{86}
\DeclareMathSymbol{\vardiamond}{\mathalpha}{extraup}{87}
\DeclareMathSymbol{\varclubsuit}{\mathalpha}{extraup}{88}

\begin{document}

\journaltitle{}
\DOI{}
\copyrightyear{}
\pubyear{}
\access{}
\appnotes{\quad}

\firstpage{1}

\title[Evaluation of Coding Schemes for Transformer-based Gene Sequence Modeling]{Evaluation of Coding Schemes for Transformer-based Gene Sequence Modeling}

\author[$^{\spadesuit*}$]{Chenlei Gong}
\author[$^{\varheart*}$]{Yuanhe Tian}
\author[$^{\clubsuit}$]{Lei Mao}
\author[$^\spadesuit$]{Yan Song}

\address[$^\spadesuit$]{University of Science and Technology of China}
\address[$^{\varheart*}$]{University of Washington}
\address[$^{\clubsuit}$]{Origin Omics}

\corresp[$^\spadesuit$]{{gongcl@mail.ustc.edu.cn}}
\corresp[$^{\varheart}$]{{yhtian@uw.edu}}
\corresp[$^{\clubsuit}$]{{maolei@originomics-ai.com}}
\corresp[$^\spadesuit$]{{clksong@gmail.com}\\$^*$Equal Contribution}

\abstract{
Currently, many studies view DNA sequences as a special type of language and utilize Transformers to model them.
These studies use fixed-length k-mer segmentation and BPE subword tokenization but lack a systematic evaluation to determine which is superior.
We compare k-mer segmentation with k=1,3,4,5,6, a 4,096-token BPE vocabulary, and three positional encoding methods—sinusoidal, AliBi, and RoPE.
Each configuration is trained from scratch in 3, 6, 12, and 24-layer Transformer encoders and evaluated on GUE benchmark dataset.
In general, BPE delivers higher and more stable performance across tasks by compressing frequent motifs into variable-length tokens, reducing sequence length, and improving model generalization.
RoPE excels at capturing periodic motifs and extrapolating to long sequences, while AliBi also performs well on tasks driven by local dependencies.
In terms of depth, we observe significant gains when increasing layers from 3 to 12, with only marginal improvements or slight overfitting at 24 layers.
This study provides practical guidance for designing tokenization and positional encoding in DNA Transformer models.\thanks{The code is available at \url{https://github.com/synlp/DNA-coding}.}
}

\maketitle

\section{Introduction}

Genomic DNA sequences are viewed through the lens of natural language, where regulatory motifs and patterns function similarly to words and syntax \citep{ji_linguistics_1999,yoon_gene_2002,searls_language_2002}.
This analogy motivates the application of language modeling techniques to DNA, using advanced deep learning models (such as Transformers \citep{vaswani2017attention}), to capture the sequence’s complex dependencies \citep{ji2021dnabert,nguyen2023hyenadna,liu2024exploring,schiff2024caduceus,nguyen2024sequence,dalla2024nucleotide,mao2025dnazen}.
These models present promising results on various DNA sequence modeling tasks in genomics, suggesting that treating DNA as a special type of language offers a powerful paradigm for understanding genomic information.

However, applying natural language processing (NLP) models to DNA sequences faces unique challenges, particularly in encoding the sequence for input to a Transformer-based model \citep{ji2021dnabert,zhang2023dnagpt,zhou2023dnabert,li_applications_2023,rozowsky_en-tex_2023,choi_transformer_2023,mao2025dnazen}.
Specifically, in sequence modeling, a token is the fundamental unit that a model processes.
In NLP, tokens at different granularities (e.g., characters, subwords, words) carry distinct semantic information and affect model performance.
Similarly, for DNA sequence modeling, an appropriate tokenization strategy is required to segment sequences into basic units.
DNA lacks clear token boundaries and instead forms a continuous string of nucleotides, making DNA sequence tokenization a non-trivial task.
A common approach segments the sequence into fixed-length k-mers as tokens \citep{wen2014k}, which captures local context but introduces information leakage and inefficiencies.
Furthermore, fixed-length k-mers may not align with biological units that vary in length.
An alternative is to learn a vocabulary of subword tokens from data, similar to byte pair encoding (BPE) \citep{sennrich2015neural} in NLP.
BPE merges frequent subsequences into variable-length tokens, which potentially capture recurring motifs of variable lengths \citep{zhou2023dnabert}.
Although these tokenization approaches achieve promising performance, it is still unclear which token units prove most effective.
In addition, encoding positional information in DNA sequence models presents a significant challenge.
Sequence order is crucial for genomic function, for example, the distance between motifs determines regulatory outcomes \citep{sharon2012inferring,jolma2015dna}.
Without explicit positional embeddings, Transformers remain agnostic to sequence order.
Standard NLP Transformers produce positional embeddings via sinusoidal functions, mapping each token position to a periodic embedding that enables the model to differentiate order.
However, this approach lacks trainable flexibility and exhibits limited extrapolation when handling sequences beyond the training length.
To better capture long-range dependencies, relative positional embeddings were introduced, injecting learnable distance-based biases into attention scores to enhance model flexibility.
Attention with linear biases (AliBi) \citep{press2021train} simplifies relative encoding by replacing learnable biases with fixed linear distance terms, enabling efficient inference and seamless support for arbitrary sequence lengths.
Rotary position embeddings (RoPE) \citep{su2024roformer} apply position-dependent rotations to query and key vectors, fusing absolute and relative positional information in dot-product attention to improve periodic pattern recognition and extrapolation capability.
These advanced positional encoding strategies are widely adopted in LLMs and genomic Transformer architectures, improving sequence modeling performance \citep{devlin2019bert,ji2021dnabert,zhang2023dnagpt,zhou2023dnabert,dalla2024nucleotide,mao2025dnazen,yang2025qwen3}.
However, given the diverse sequence lengths and genomic architectures across tasks, no consensus emerges.
Therefore, it is essential to explore which positional encoding type works for DNA sequence modeling with Transformer-based models.

\begin{figure*}[t]
\centering
\includegraphics[width=1\textwidth]{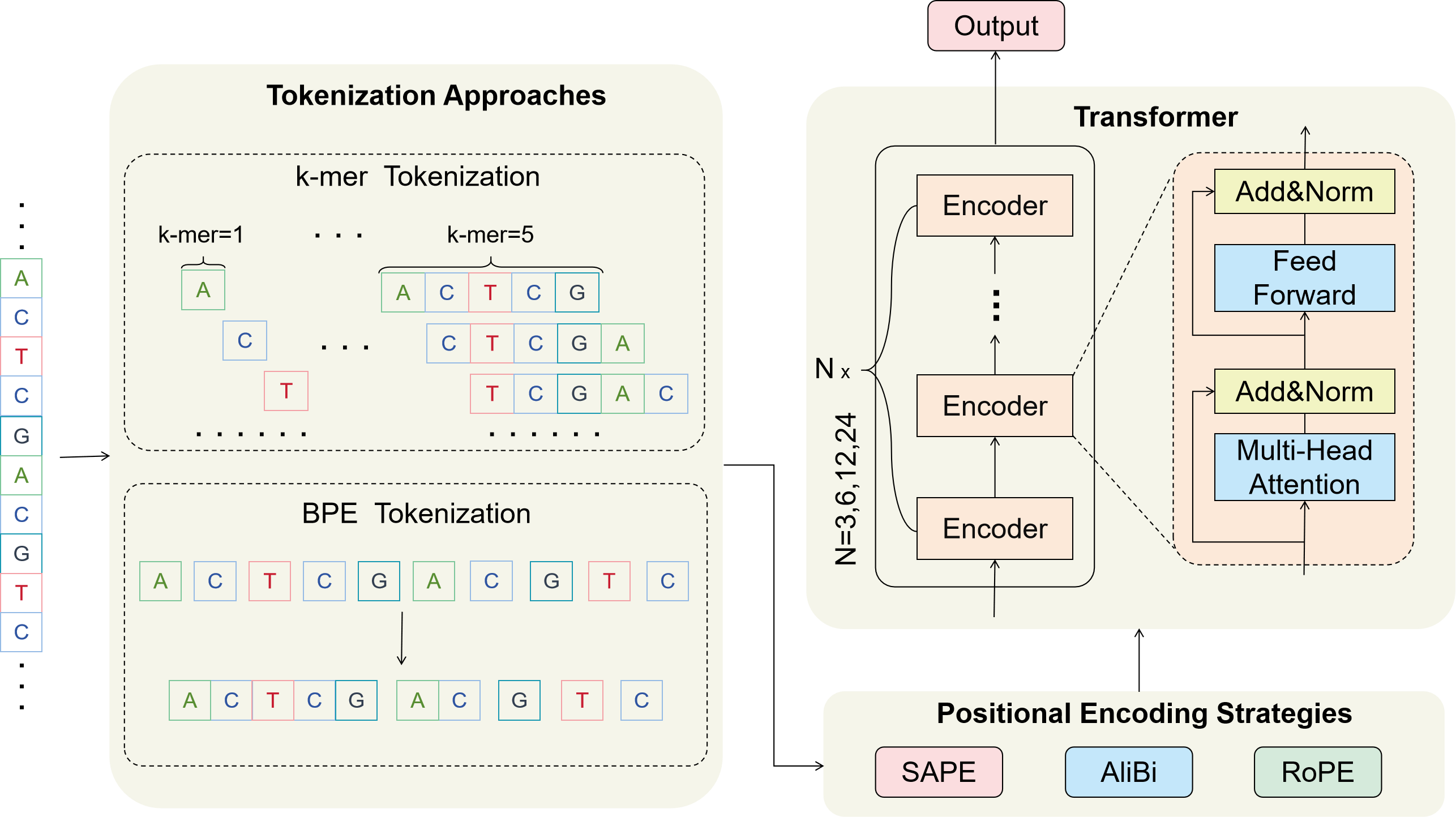}
\caption{The overall architecture of our approach.}
\label{fig:architecture}
\end{figure*}

This paper systematically evaluates DNA sequence encoding strategies based on the Transformer architecture.
We model DNA as a symbolic language and explore encoding methods at two levels.
First, we compare different DNA sequence segmentation methods, including traditional continuous k-mer tokenization and BPE-style vocabulary learning.
Second, we compare multiple positional embedding methods for injecting sequence position information into Transformers, including sinusoidal absolute positional encoding (SAPE), attention with linear biases (AliBi), and rotary position embeddings (RoPE).
Additionally, we investigate the impact of varying Transformer depths.
We evaluate the performance of different models on six representative genomic classification tasks from the GUE benchmark dataset.
These tasks include Core Promoter Detection, Transcription Factor Prediction, Promoter Detection, Splice Site Prediction, Epigenetic Mark Prediction, and COVID Variant Classification.
They cover different species, sequence lengths, and prediction objectives.
We find that BPE tokenization outperforms fixed-length k-mers in most tasks, and k-mer encodings with different k values exhibit varying performance across tasks.
We observe that Rotary positional embeddings significantly outperform other positional encoding schemes in DNA modeling, particularly in extrapolation ability and capturing periodic patterns.
We observe that larger-scale Transformer models improve performance on most tasks, though with diminishing returns.
These findings provide guidance for future model design.

\section{Related Work}

\subsection{DNA Sequence Tokenization}

DNA sequence tokenization is a fundamental preprocessing step that determines the atomic units for downstream genomic modeling.
An intuitive approach is to use single nucleotides, i.e., A, C, G, T, as tokens, but this minimal vocabulary limits contextual representation and yields very long token sequences  \citep{zhou2015predicting,kelley2016basset}.
To address these issues, fixed-length k-mer segmentation extracts overlapping substrings of length k as tokens \citep{wen2014k}, capturing local context but suffering from exponential vocabulary growth and rigid granularity \citep{liang2020deepmicrobes,mock2022taxonomic,ji2021dnabert,sanabria2024dna,zhou2015predicting}.
Inspired by NLP, subword tokenization methods such as Byte Pair Encoding (BPE) learn variable-length tokens by merging frequent nucleotide pairs \citep{sennrich2015neural}, aligning more naturally with biological motifs and reducing sequence length.  
Some studies explore hybrid schemes that combine k-mer and subword units to capture multi-scale patterns \citep{wu2025generator}, further improving efficiency and accuracy.
Despite these advances, existing work lacks a systematic comparison of tokenization strategies in transformer-based DNA sequence models.

\subsection{Position Encoding of Transformer}

Position encoding strategies, which inject order information into Transformer models, are both varied and crucial.
Early Transformers use either learned absolute position embeddings or sinusoidal functions to encode absolute positional information \citep{vaswani2017attention,devlin2019bert}.
For example, BERT \citep{devlin2019bert} adopts fixed-length learned absolute embeddings, restricting the input to 512 nucleotides.
Transformer-XL \citep{dai2019transformer} introduces relative position bias, using learnable distance-dependent biases to handle variable-length contexts.
T5 \citep{raffel2020exploring} buckets relative distances to reduce parameter cost while retaining fine-grained bias information.
ALiBi \citep{press2021train} replaces learnable biases with fixed linear distance decay terms, achieving simplicity, efficiency, and seamless support for arbitrary lengths.
RoPE \citep{su2024roformer} applies position-dependent rotations to query and key vectors, fusing absolute and relative information with strong extrapolation capability.
Other general approaches include axial positional encoding for dimensional decomposition and differentiable relative encoding to improve cross-layer dependency representation \citep{kitaev2020reformer,he2020deberta,liutkus2021relative}.
Despite the variety of schemes, a systematic study of their combined effects and applicability boundaries in genomic tasks remains lacking.

\section{Approach}

In our approach, we apply two tokenization schemes to DNA sequences: fixed-length k-mer segmentation and data-driven BPE-based subword tokenization.
We compare three positional encoding techniques: learned absolute embeddings, relative position bias, and rotary position embeddings.
We use a multi-layer Transformer encoder to integrate token and positional information and evaluate on downstream DNA sequence modeling tasks.
The overall architecture is illustrated in Figure \ref{fig:architecture}.
The details are illustrated as follows.

\subsection{Tokenization Approaches}

We considered two tokenization schemes for representing DNA sequences as model inputs: fixed-length contiguous $k$-mers and Byte Pair Encoding (BPE) subwords.

\subsubsection{$k$-mer Tokenization}

k-mer is a method that extracts substrings from DNA sequences using a sliding window of length k with step size 1.
This method treats every contiguous substring of length k in the sequence as a token.
For different values of k, the vocabulary size equals $4^k$.
A sequence of length N generates $N-k+1$ tokens under k-mer encoding.
Common k values include 3, 4, and 6, corresponding to vocabulary sizes of 64, 256, and 4096, respectively.
In particular, when k is 1, the nucleobase is directly used as the token.
For example, a sequence ATGCGTACG of length 9 generates 7 k-mers at k=3: [ATG, TGC, GCG, CGT, GTA, TAC, ACG].
Overall, k-mer tokenization approach is simple and enumerates all k-length substrings.
The vocabulary size grows exponentially as $4^k$, resulting in high computation and memory costs when k is large.
In addition, k-mer captures only single-scale motifs and cannot adaptively represent multi-scale biological signals.

\subsubsection{Byte Pair Encoding (BPE) Subword Tokenization}

Byte pair encoding (BPE) is a greedy algorithm that learns variable-length subword units by iteratively merging the most frequent pair of symbols in the corpus.
The algorithm starts from an initial alphabet (A, C, G, T) and merges the most frequent adjacent symbol pair at each iteration.
Mathematically, the symbol pair merged at iteration t satisfies:
\begin{equation}
(a, b) = \arg\max_{x,y} \mathrm{count}(xy)
\end{equation}
After each merge, the vocabulary size increases by one and all corresponding instances in the corpus are replaced by the new symbol.
By performing T merge iterations, the resulting vocabulary size becomes approximately $|V_0| + T$.
Each vocabulary contains single-nucleotide symbols as well as multi-nucleotide subwords to capture genome motifs at various scales.
Overall, BPE learns subwords from frequency statistics and effectively compresses frequent sequence motifs.
It captures multi-scale variable-length biological patterns and reduces the token count.

\subsection{Positional Encoding Strategies}

Positional encoding injects order information into Transformers’ inherently order-agnostic self-attention mechanism.
We evaluate three positional encoding strategies: standard sinusoidal position embeddings, attention with linear biases (AliBi), and rotary position embeddings (RoPE).
Each method offers distinct strengths in order preservation, extrapolation capability, and representational characteristics.
The following sections detail the principles and implementations of these mechanisms.

\subsubsection{Sinusoidal Absolute Position Embeddings}

Sinusoidal absolute position embeddings (SAPE) inject order information by applying sine and cosine functions at different frequencies for each position pos.
For position pos and dimension index i, the encoding is defined as:
\begin{equation}
\mathrm{PE}(pos,2i) = \sin\bigl(\tfrac{pos}{10000^{2i/d}}\bigr)
\end{equation}
and
\begin{equation}
\mathrm{PE}(pos,2i+1) = \cos\bigl(\tfrac{pos}{10000^{2i/d}}\bigr)
\end{equation}
At input, the positional vector is added to the corresponding token embedding to obtain the final representation.
This method requires no trainable parameters and maintains consistent encoding patterns beyond the training length.

\begin{table*}[t]
\centering
\caption{Summary of tasks and statistics in the GUE benchmarks.}
\scalebox{1.0}{
\begin{tabular}{lccccccc}
\toprule
Species & Task & Num. Datasets & Num. Classes & Sequence Length & train.N.s & dev.N.s & test.N.s \\
\midrule
Human & Core Promoter Detection & 3 & 2 & 70 & 94712 & 11840 & 11840\\
Human & Transcription Factor Prediction & 5 & 2 & 100 & 128345 & 5000 & 5000 \\
Human & Promoter Detection & 3 & 2 & 300 & 93902 & 11840 & 11840\\
Human & Splice Site Prediction & 1 & 3 & 400 & 36496 & 4562 & 4562 \\
Mouse & Transcription Factor Prediction & 5 & 2 & 100 & 80018 & 9735 & 9735\\
Yeast & Epigenetic Marks Prediction & 10 & 2 & 500 & 229885 & 28741 & 28741\\
Virus & Covid Variant Classification & 1 & 9 & 1000 & 73335 & 9168 & 9166 \\
\bottomrule
\end{tabular}
}
\label{tab:tasks}
\end{table*}

\begin{table*}[tp]
\centering
\caption{The results of models using absolute position encoding strategies.}
\label{tab:abs}
\begin{tabular}{c|c|ccccccc}
\toprule
Layers & Tokenization & Human-CPD & Human-PD & Human-SSD & Human-FTP & Mouse-TFP & Virus-Covid & Yest-EMP \\
\midrule
\multirow{6}{*}{3}
& 1-mer  & 0.4769 & 0.4584 & 0.4910 & 0.3573 & 0.1928 & 0.1012 & 0.1969 \\
& 3-mer  & 0.5435 & 0.5848 & 0.5017 & 0.5004 & 0.3330 & 0.4132 & 0.3614 \\
& 4-mer  & 0.5534 & 0.5627 & 0.5029 & 0.5112 & 0.4703 & 0.4021 & 0.3846 \\
& 5-mer  & 0.5464 & 0.5824 & 0.5079 & 0.5267 & 0.4829 & 0.4206 & 0.3922 \\
& 6-mer  & 0.5393 & 0.5735 & 0.5141 & 0.5134 & 0.3786 & 0.4579 & 0.3451 \\
& BPE    & 0.5636 & 0.6028 & 0.5559 & 0.5394 & 0.5026 & 0.6169 & 0.4048 \\
\midrule
\multirow{6}{*}{6} 
& 1-mer  & 0.4869 & 0.5193 & 0.4833 & 0.3685 & 0.1430 & 0.1073 & 0.2031 \\
& 3-mer  & 0.5313 & 0.5962 & 0.5210 & 0.5124 & 0.3819 & 0.4510 & 0.3765 \\
& 4-mer  & 0.5459 & 0.5741 & 0.5027 & 0.5829 & 0.4875 & 0.4496 & 0.3985 \\
& 5-mer  & 0.5724 & 0.5928 & 0.5405 & 0.6000 & 0.5423 & 0.4163 & 0.4077 \\
& 6-mer  & 0.5392 & 0.6026 & 0.5380 & 0.5993 & 0.4309 & 0.4611 & 0.4075 \\
& BPE    & 0.5587 & 0.6059 & 0.5309 & 0.5262 & 0.4970 & 0.6493 & 0.4049 \\
\midrule
\multirow{6}{*}{12} 
& 1-mer  & 0.4637 & 0.4660 & 0.4508 & 0.4032 & 0.2162 & 0.1109 & 0.2171 \\
& 3-mer  & 0.5542 & 0.6041 & 0.5436 & 0.5244 & 0.3727 & 0.4242 & 0.3811 \\
& 4-mer  & 0.5473 & 0.5824 & 0.5247 & 0.5720 & 0.4834 & 0.4395 & 0.3914 \\
& 5-mer  & 0.5347 & 0.6200 & 0.5065 & 0.6023 & 0.5624 & 0.4097 & 0.4097 \\
& 6-mer  & 0.5542 & 0.5721 & 0.4552 & 0.5997 & 0.4504 & 0.4286 & 0.4173 \\
& BPE    & 0.5555 & 0.6239 & 0.5309 & 0.5256 & 0.4995 & 0.6556 & 0.4030 \\
\midrule
\multirow{6}{*}{24} 
& 1-mer  & 0.4640 & 0.4313 & 0.4675 & 0.4286 & 0.3017 & 0.1165 & 0.2013 \\
& 3-mer  & 0.5208 & 0.6059 & 0.4862 & 0.5178 & 0.3185 & 0.4562 & 0.3807 \\
& 4-mer  & 0.5013 & 0.5816 & 0.5500 & 0.5540 & 0.3945 & 0.4614 & 0.3852 \\
& 5-mer  & 0.5329 & 0.5586 & 0.5362 & 0.5603 & 0.3700 & 0.4137 & 0.3521 \\
& 6-mer  & 0.5065 & 0.5828 & 0.5021 & 0.5471 & 0.4023 & 0.4384 & 0.3454 \\
& BPE    & 0.5409 & 0.6377 & 0.4991 & 0.5290 & 0.4339 & 0.6528 & 0.3965 \\
\bottomrule
\end{tabular}
\end{table*}

\begin{table*}[tp]
\centering
\caption{The results of models using AliBi position encoding strategies.}
\label{tab:alibi}
\begin{tabular}{c|c|ccccccc}
\toprule
Layers & Tokenization & Human-CPD & Human-PD & Human-SSD & Human-FTP & Mouse-TFP & Virus-Covid & Yest-EMP \\
\midrule
\multirow{6}{*}{3}
& 1-mer  & 0.5274 & 0.5130 & 0.4724 & 0.3176 & 0.2141 & 0.1056 & 0.2613 \\
& 3-mer  & 0.6076 & 0.5649 & 0.5143 & 0.5254 & 0.3738 & 0.4461 & 0.3696 \\
& 4-mer  & 0.5749 & 0.5628 & 0.4938 & 0.5671 & 0.4271 & 0.4349 & 0.3334 \\
& 5-mer  & 0.5771 & 0.5319 & 0.4821 & 0.5914 & 0.4667 & 0.4412 & 0.3028 \\
& 6-mer  & 0.5451 & 0.5041 & 0.5241 & 0.6011 & 0.4728 & 0.4460 & 0.2702 \\
& BPE    & 0.5445 & 0.6445 & 0.5611 & 0.5414 & 0.4734 & 0.6681 & 0.3777 \\
\midrule
\multirow{6}{*}{6} 
& 1-mer  & 0.4914 & 0.5545 & 0.4729 & 0.3923 & 0.2346 & 0.1121 & 0.2845 \\
& 3-mer  & 0.5999 & 0.5984 & 0.5091 & 0.5545 & 0.4209 & 0.4567 & 0.3991 \\
& 4-mer  & 0.6123 & 0.6062 & 0.5154 & 0.5954 & 0.4762 & 0.4568 & 0.3728 \\
& 5-mer  & 0.5790 & 0.5967 & 0.5143 & 0.6175 & 0.5325 & 0.4303 & 0.3538 \\
& 6-mer  & 0.5568 & 0.5424 & 0.5021 & 0.6206 & 0.5168 & 0.4585 & 0.2994 \\
& BPE    & 0.5394 & 0.6582 & 0.5355 & 0.5359 & 0.4894 & 0.6788 & 0.3990 \\
\midrule
\multirow{6}{*}{12} 
& 1-mer  & 0.5099 & 0.6261 & 0.4771 & 0.4326 & 0.2555 & 0.1280 & 0.3389 \\
& 3-mer  & 0.6161 & 0.6187 & 0.5253 & 0.5722 & 0.4517 & 0.4682 & 0.4112 \\
& 4-mer  & 0.6207 & 0.6012 & 0.5371 & 0.6004 & 0.5182 & 0.4559 & 0.4048 \\
& 5-mer  & 0.6065 & 0.6016 & 0.5327 & 0.6205 & 0.5671 & 0.4558 & 0.3909 \\
& 6-mer  & 0.5536 & 0.5690 & 0.5119 & 0.6291 & 0.5037 & 0.4607 & 0.3403 \\
& BPE    & 0.5503 & 0.6471 & 0.5565 & 0.5381 & 0.5044 & 0.6644 & 0.4066 \\
\midrule
\multirow{6}{*}{24} 
& 1-mer  & 0.5089 & 0.6202 & 0.4821 & 0.4694 & 0.3628 & 0.1372 & 0.3324 \\
& 3-mer  & 0.5053 & 0.6011 & 0.5032 & 0.5291 & 0.4324 & 0.4138 & 0.4038 \\
& 4-mer  & 0.5132 & 0.6126 & 0.5749 & 0.5362 & 0.4862 & 0.4546 & 0.3923 \\
& 5-mer  & 0.5239 & 0.5736 & 0.5363 & 0.5521 & 0.4723 & 0.4470 & 0.3719 \\
& 6-mer  & 0.5382 & 0.5697 & 0.5664 & 0.5652 & 0.4821 & 0.4329 & 0.3273 \\
& BPE    & 0.5347 & 0.6844 & 0.5824 & 0.5509 & 0.4977 & 0.6867 & 0.3903 \\
\bottomrule
\end{tabular}
\end{table*}

\begin{table*}[tp]
\centering
\caption{The results of models using Rotary position encoding strategies.}
\label{tab:rotary}
\begin{tabular}{c|c|ccccccc}
\toprule
Layers & Tokenization & Human-CPD & Human-PD & Human-SSD & Human-FTP & Mouse-TFP & Virus-Covid & Yest-EMP \\
\midrule
\multirow{6}{*}{3}
& 1-mer  & 0.4979 & 0.5629 & 0.5064 & 0.3857 & 0.3843 & 0.1049 & 0.2074 \\
& 3-mer  & 0.5154 & 0.5873 & 0.5336 & 0.5121 & 0.4234 & 0.3458 & 0.3481 \\
& 4-mer  & 0.5624 & 0.5701 & 0.5439 & 0.5784 & 0.5003 & 0.3954 & 0.3903 \\
& 5-mer  & 0.5676 & 0.5184 & 0.5469 & 0.5133 & 0.4693 & 0.4155 & 0.3351 \\
& 6-mer  & 0.5206 & 0.5042 & 0.5384 & 0.5124 & 0.5016 & 0.4579 & 0.3121 \\
& BPE    & 0.5527 & 0.6648 & 0.5733 & 0.5492 & 0.4953 & 0.6284 & 0.4017 \\
\midrule
\multirow{6}{*}{6} 
& 1-mer  & 0.5027 & 0.5827 & 0.5182 & 0.4035 & 0.4026 & 0.1181 & 0.2243 \\
& 3-mer  & 0.5329 & 0.5917 & 0.5427 & 0.4561 & 0.4439 & 0.2743 & 0.3246 \\
& 4-mer  & 0.5654 & 0.5748 & 0.5627 & 0.5898 & 0.5224 & 0.3942 & 0.3890 \\
& 5-mer  & 0.5458 & 0.5253 & 0.5440 & 0.5438 & 0.4184 & 0.4478 & 0.3626 \\
& 6-mer  & 0.5445 & 0.5727 & 0.5439 & 0.5256 & 0.5258 & 0.4611 & 0.3455 \\
& BPE    & 0.5636 & 0.6736 & 0.5822 & 0.5546 & 0.5190 & 0.6417 & 0.4056 \\
\midrule
\multirow{6}{*}{12} 
& 1-mer  & 0.5212 & 0.6471 & 0.5391 & 0.4428 & 0.4125 & 0.1357 & 0.2392 \\
& 3-mer  & 0.5514 & 0.5859 & 0.5649 & 0.4443 & 0.4302 & 0.4468 & 0.3681 \\
& 4-mer  & 0.5732 & 0.5761 & 0.5729 & 0.5547 & 0.4728 & 0.4174 & 0.3609 \\
& 5-mer  & 0.5674 & 0.5641 & 0.5542 & 0.5910 & 0.4816 & 0.4308 & 0.3779 \\
& 6-mer  & 0.5495 & 0.5993 & 0.5634 & 0.6086 & 0.5017 & 0.4590 & 0.3613 \\
& BPE    & 0.5588 & 0.6642 & 0.5800 & 0.5500 & 0.5096 & 0.6535 & 0.4189 \\
\midrule
\multirow{6}{*}{24} 
& 1-mer  & 0.5012 & 0.6680 & 0.5448 & 0.4916 & 0.4368 & 0.1543 & 0.2380 \\
& 3-mer  & 0.5239 & 0.6028 & 0.5599 & 0.5274 & 0.4395 & 0.4615 & 0.3872 \\
& 4-mer  & 0.5103 & 0.5844 & 0.5865 & 0.5310 & 0.4523 & 0.4733 & 0.3778 \\
& 5-mer  & 0.5305 & 0.5652 & 0.5732 & 0.5459 & 0.4726 & 0.4856 & 0.3633 \\
& 6-mer  & 0.5542 & 0.5941 & 0.5853 & 0.5626 & 0.4949 & 0.4732 & 0.3706 \\
& BPE    & 0.5672 & 0.6791 & 0.6064 & 0.5548 & 0.5195 & 0.6850 & 0.4107 \\
\bottomrule
\end{tabular}
\end{table*}

\subsubsection{Attention with Linear Biases (AliBi)}

Attention with Linear Biases (AliBi) adds a linear bias term that scales with token distance into self-attention computations.
For head \(h\), the attention score between positions \(i\) and \(j\) is computed as:
\begin{equation}
\text{score}_{h}(i,j) \;=\; \frac{\mathbf{q}_{i}^{h}\cdot\mathbf{k}_{j}^{h}}{\sqrt{d}} \;+\; m_{h}\,\lvert i - j\rvert
\end{equation}
Herein, the slope parameter \(m_{h}\) is a fixed hyperparameter chosen empirically and remains constant during training.
This method does not require additional trainable position embeddings and generalizes seamlessly to any sequence length.
It emphasizes local dependencies while preserving computational efficiency.

\subsubsection{Rotary Position Embeddings (RoPE)}

Rotary Position Embeddings (RoPE) applies a two-dimensional rotation to each adjacent pair of dimensions in query and key vectors based on token position.
For position pos and dimension pair index i, the rotation angle is defined as:
\begin{equation}
\theta_{pos,i} = \frac{pos}{10000^{2i/d}}
\end{equation}
Each subvector \(\bigl[x_{2i}, x_{2i+1}\bigr]^T\) is transformed as
\begin{equation}
\mathrm{Rot}\bigl([x_{2i}, x_{2i+1}]^T,\,pos\bigr) =
\begin{pmatrix}
\cos\theta_{pos,i} & -\sin\theta_{pos,i}\\
\sin\theta_{pos,i} & \cos\theta_{pos,i}
\end{pmatrix}
\begin{pmatrix}
x_{2i}\\ x_{2i+1}
\end{pmatrix}
\end{equation}
where the dot product of rotated \(q_i\) and \(k_j\) naturally encodes a periodic function of \((pos_i - pos_j)\).
This approach preserves vector norms and requires no additional trainable parameters, generalizing seamlessly to any sequence length.

\subsection{Transformer-based DNA Sequence Classification}

Transformer is widely used across various domains and has become a mainstream architecture for sequence encoding \citep{devlin2019bert,tian-etal-2021-federated,qin-2022-complementary,ouyang2022training,touvron2023llama}.
To validate the effectiveness of different encoding strategies, we employ a Transformer model for DNA sequence classification.
Given an input x, we first segment it into a token sequence using the aforementioned tokenization methods.
We then append special start and end tokens to the token sequence.
Next, we map each token to a d-dimensional embedding, forming the embedding matrix \(\mathbf{E}\).
The embedding matrix is encoded by a Transformer encoder composed of \(l\) layers.
The encoder layer \(k\) computes \(\mathbf{H}^{(k)}\) based on the output of the previous layer:
\begin{equation}
\mathbf{H}^{(k)} = \text{Layer}_{k}\bigl(\mathbf{H}^{(k-1)}\bigr)
\end{equation}
where $k=1 \dots l$ and $\mathbf{H}^{(0)} = \mathbf{E}$.
We extract \(\mathbf{H}^{(l)}_{0}\) (i.e., the representation of the initial special token) as the global feature.
Finally, we feed this feature into a fully connected layer to obtain the classification prediction \(\hat{\mathbf{y}}\).

\section{Experiment Setting}
\label{sec:exp-setup}

\subsection{Datasets and Tasks}
\label{sec:datasets}

We evaluate models with various settings on the Genome Understanding Evaluation (GUE) benchmark \citep{zhou2023dnabert}, which contains six representative genomic sequence classification tasks:
\begin{itemize}
\item \textbf{Core Promoter Detection (CPD)}: identify whether a DNA sequence contains core promoter regions.
  \item \textbf{Transcription Factor Prediction (TFP)}: predict whether a DNA sequence contains binding sites of transcription factors.
  \item \textbf{Promoter Detection (PD)}: identify whether a DNA sequence contains general promoter region.
  \item \textbf{Splice Site Prediction (SSP)}: distinguish true splice junctions from decoys.
  \item \textbf{Epigenetic Mark Prediction (EMP)}: predict presence of histone modifications or DNA methylation.
  \item \textbf{COVID Variant Classification (CVC)}: classify SARS-CoV-2 spike-protein sequences by variant.
\end{itemize}
For each task, we use the standard GUE train, development, and test splits.
The statistics of the data and tasks are reported in Table \ref{tab:tasks}.

\subsection{Implementation Details}
\label{sec:impl-details}

In the experiments, we try different configurations of the tokenization strategies, positional encoding strategies, and model sizes.
For the tokenization approaches, we compare two representative ones: k-mer tokenization (with k=1,3,4,6) and subword tokenization based on Byte Pair Encoding (BPE).
We use the BPE tokenizer used in previous studies \citep{zhou2023dnabert,mao2025dnazen}, where there are 4,096 DNA tokens in the vocabulary.
To explore the influence of positional encoding methods, we evaluated three mainstream approaches: sinusoidal absolute positional encoding (SAPE), attention with linear biases (AliBi), and rotary position embeddings (RoPE).
For the model, we employ a Transformer encoder \citep{vaswani2017attention} architecture to process DNA sequence inputs.
The number of self-attention layers in the Transformer encoder is set to 3, 6, 12, and 24, where the dimension of the hidden vector is set to 768 for all variants, which leads to 32M, 60M, 117M, and 230M number of parameters, respectively.
In training, we use the AdamW \citep{loshchilov2017decoupled} optimizer with a learning rate of $1×10^{-4}$ and weight decay of 0.01.
We apply a linear warmup over the first 10\% of training steps, followed by cosine decay to zero, and use a dropout rate of 0.1 on both attention weights and feed-forward outputs.
To ensure a fair comparison, all models with different settings (different tokenizers, positional encodings, and layer numbers) are trained from scratch, without using any pretrained weights, minimizing the confounding effects of prior training.
For evaluation, we adopt the Matthews correlation coefficient (MCC) as our primary performance metric, following existing studies \citep{ji2021dnabert,zhou2023dnabert,mao2025dnazen}.
MCC provides a stable measure of classification consistency under class imbalance, with higher scores indicating better model performance.

\begin{figure}[tp]
\centering
\includegraphics[width=0.5\textwidth, trim=0 20 0 0]{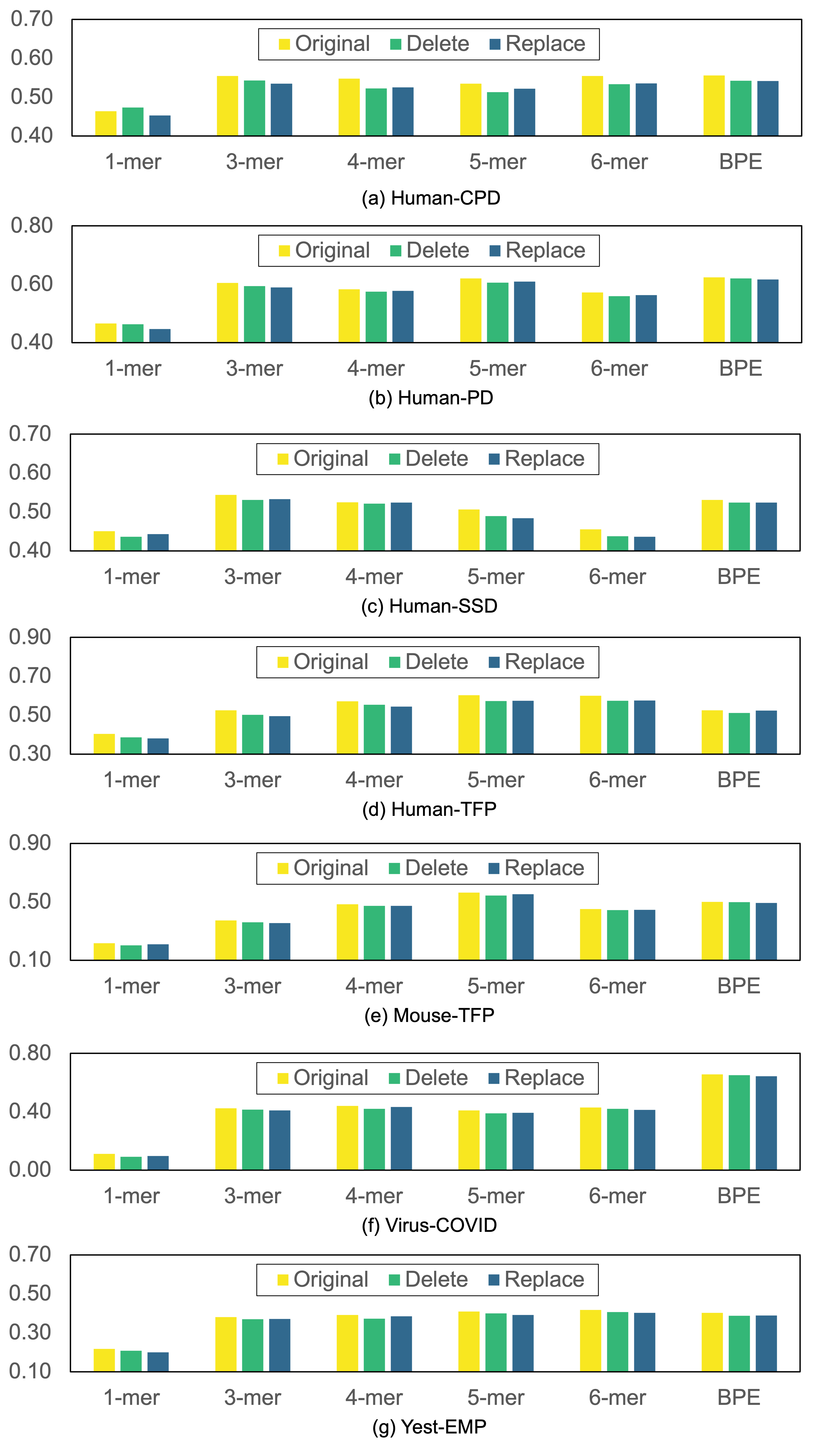}
\caption{The results of models (using 12 layers) with different settings when using the absolute position encoding.}
\label{fig:e-results1}
\end{figure}

\begin{figure}[tp]
\centering
\includegraphics[width=0.5\textwidth, trim=0 20 0 0]{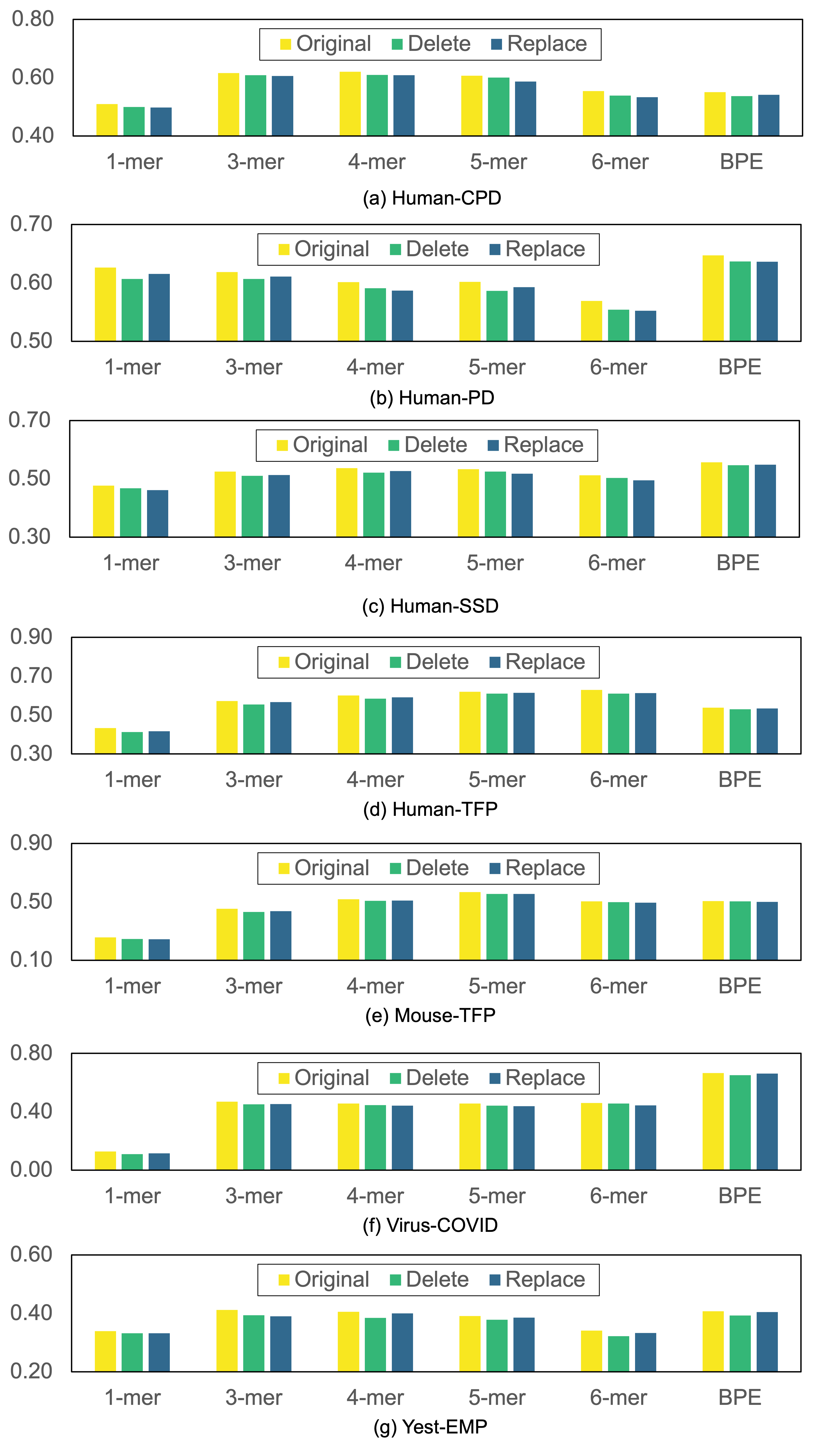}
\caption{The results of models (using 12 layers) with different settings when using the AliBi position encoding.}
\label{fig:e-results2}
\end{figure}

\begin{figure}[tp]
\centering
\includegraphics[width=0.5\textwidth, trim=0 20 0 0]{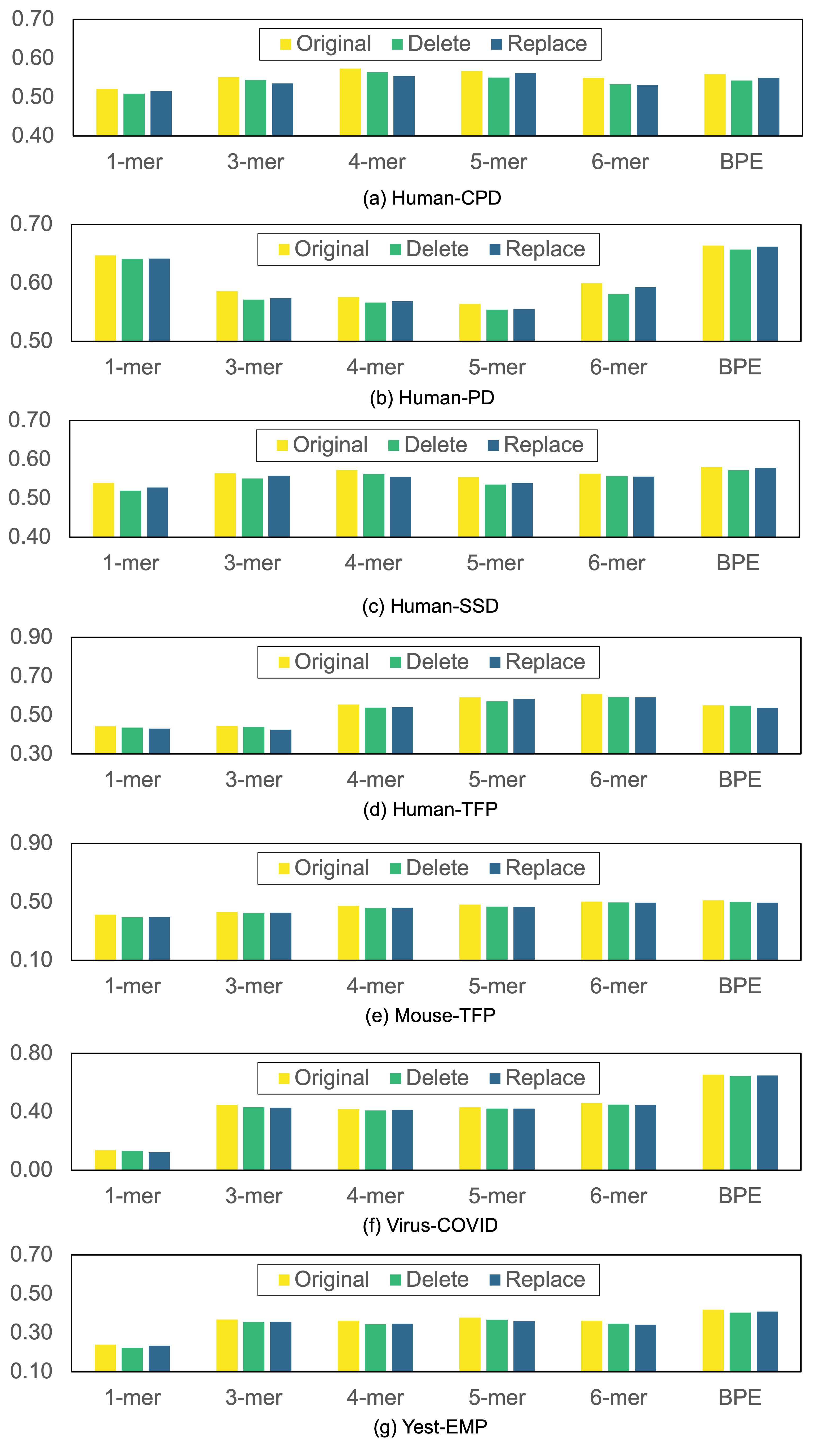}
\caption{The results of models (using 12 layers) with different settings when using the Rotary position encoding.}
\label{fig:e-results3}
\end{figure}

\section{Results and Analysis}

Our main experiment results of using absolute position encoding, AliBi, and RoPE are presented in Table \ref{tab:abs}, Table \ref{tab:alibi}, and Table \ref{tab:rotary}, respectively, where the performance on each task is the average performance on all datasets of the task.
Each Table shows the results of different tokenization strategies with various numbers of Transformer layers.
Based on these results, in the following content, we analyze the impacts of different tokenization approaches, different positional encoding methods, and different numbers of Transformer layers.

\subsection{Impact of Tokenization Approach}

Overall, BPE demonstrates the best performance across different experiment settings compared with k-mer tokenization approaches with various k values, demonstrating the strong ability of BPE to capture various data features.
This is because BPE is able to capture common biological motifs with variable-length subwords and effectively compresses sequence length.
Because the BPE vocabulary includes both single-nucleotide and multi-nucleotide substrings, it balances local and global information, enhancing the model’s ability to represent multi-scale signals.
Among different settings of the k-mer tokenization strategies, the case when $k=1$ (i.e., using a single nucleotide unit as a token) achieves relatively low performance compared with other settings.
This may be because the 1-mer extracts overly simple features that fail to fully capture complex patterns in the data.
As k increases from 3 to 6, performance diverges across tasks, indicating that the optimal k value is closely related to the typical length of biological signals.
For regulatory signals based on trinucleotides (such as splice sites), k=3 often yields better recognition, whereas larger k values perform better in epigenetic mark prediction tasks that require detection of longer motifs.
However, overly large k values (e.g., k=6) lead to exploding vocabulary size and increased sparsity, significantly raising training and inference costs and introducing potential overfitting in some tasks.
These results suggest that the optimal fixed-length k-mer setting must align with the signal scale of a specific task, whereas BPE offers a more adaptive tokenization scheme.

In bioinformatics, DNA sequences often undergo minor changes due to factors such as extraction start point shifts or mutations, which may degrade the performance of pretrained models on test data.
To systematically assess the impact of nucleotide alterations on model robustness, we evaluate pre-trained models under three settings: (1) the original sequence set, (2) the group with random substitutions at both ends, and (3) the group with the first three bases deleted and randomly filled at the tail.
We report the results of models (with 12 Transformer layers) with absolute position encoding, AliBi, and RoPE in Figure \ref{fig:e-results1}, Figure \ref{fig:e-results2}, and Figure \ref{fig:e-results3}, respectively.
Experiment results show that every tokenization approach experiences performance degradation when faced with nucleotide changes, but the magnitude of decline is modest, indicating inherent model robustness.
Among different tokenization strategies, BPE shows the smallest performance fluctuation across the three test groups.
This advantage arises because BPE subwords capture multi-scale patterns around mutation sites, mitigating the impact of sparse nucleotide changes on overall representations.
For larger k-mer encodings, the performance drop is greater, as more k-mer tokens are entirely altered by minimal nucleotide variations.
These observations underscore the importance of adopting adaptive tokenization strategies to enhance the stability of DNA sequence models under real-world variation conditions.

\subsection{Impact of Positional Encoding Strategy}

Overall, Rotary achieves the best performance on the tasks, followed by AliBi, with standard absolute positional encoding obtaining the worst results.
By applying rotations to query and key vectors, Rotary fuses absolute and relative positional information, achieving superior extrapolation and multi-scale signal perception.
AliBi is good at emphasizing local dependencies and simplicity of implementation, but its linear distance bias lacks sensitivity to complex periodic signals.
Standard absolute positional encoding merely provides fixed global coordinate markers, with limited extrapolation to ultra-long sequences and insufficient capture of fine-grained relative relationships.

\subsection{Impact of the Number of Transformer Layers}

Overall, Transformer models with varying depths show a rapid-to-slow growth trend in performance rather than a purely increasing or decreasing pattern.
The 3-layer model achieves the lowest average MCC, likely because its shallow depth struggles to capture complex patterns and multi-scale dependencies in the data.
As the number of layers increases from 3 to 12, performance markedly improves, indicating that deeper networks can learn richer sequence relationships and periodic signals.
When extended to 24 layers, performance continues to improve but with significantly smaller gains, reflecting diminishing marginal returns from increased depth.
The diminishing performance gains may stem from increased training difficulty, higher risks of vanishing gradients, and heightened overfitting potential.
These observations suggest that when selecting model depth, one must balance representational capacity against computational cost and avoid stacking layers indiscriminately.

\section{Discussion}

The study demonstrates that the tokenization strategy influences genomic sequence model performance, particularly in capturing multi-scale patterns.
Fixed-length k-mer tokens show acceptable results for simple sequence tasks but lack sensitivity to crucial local motifs due to their uniform granularity and generate redundancy in long sequences.
In contrast, BPE subwords adaptively encode frequent motifs with variable lengths, enhancing the model’s representation of diverse signals and maintaining higher, more stable performance across tasks.
In robustness tests against minor nucleotide alterations, BPE shows the smallest performance fluctuations, indicating greater adaptability and fault tolerance to sequence tweaks.
For positional encoding, Rotary embeddings are good at capturing periodic signals and extrapolating to ultra-long sequences by integrating both absolute and relative information for rich feature representation.
AliBi effectively reinforces local dependencies with linear distance bias, thus delivering consistent performance in tasks driven by short-range interactions.
Standard absolute positional encoding provides only fixed global coordinates, offering limited support for complex relative relationships and extrapolation to ultra-long sequences.
Further analysis shows that as Transformer depth increases from 3 to 24 layers, performance steadily improves while gains progressively diminish.
These findings suggest that future genomic language model designs should jointly consider tokenization, positional encoding, and depth to strike an optimal balance between performance and efficiency.

\section{Conclusion}

In this work, we present a systematic evaluation of encoding strategies for transformer-based DNA sequence models.
We compare fixed-length k-mer and BPE tokenizations, alongside three positional encoding methods: sinusoidal absolute embeddings, attention with linear biases (AliBi), and rotary position embeddings (RoPE).
Our experiments span three model depths (3, 6, 12, and 24 layers) and several tasks from the Genome Understanding Evaluation benchmark.
Our results show that BPE subwords outperform fixed k-mer segmentation in capturing multi-scale biological patterns.
For positional encoding, Rotary position embeddings demonstrate superior extrapolation capability and periodic motif recognition.
AliBi excels at emphasizing local dependencies, and absolute embeddings perform reliably on tasks with fixed motif spacing.
We find that increasing model depth shows performance gains.
These findings underscore the importance of co-designing token granularity and positional encoding based on task requirements and provide guidance for future development of DNA Transformer models.

\nocite{langley00}

\bibliographystyle{abbrvnat}
\bibliography{reference}

\begin{thebibliography}{40}
\providecommand{\natexlab}[1]{#1}
\providecommand{\url}[1]{\texttt{#1}}
\expandafter\ifx\csname urlstyle\endcsname\relax
  \providecommand{\doi}[1]{doi: #1}\else
  \providecommand{\doi}{doi: \begingroup \urlstyle{rm}\Url}\fi

\bibitem[Choi and Lee(2023)]{choi_transformer_2023}
S.~R. Choi and M.~Lee.
\newblock Transformer architecture and attention mechanisms in genome data analysis: A comprehensive review.
\newblock 12\penalty0 (7):\penalty0 1033, 2023.
\newblock ISSN 2079-7737.
\newblock Number: 7 Publisher: Multidisciplinary Digital Publishing Institute.

\bibitem[Dai et~al.(2019)Dai, Yang, Yang, Carbonell, Le, and Salakhutdinov]{dai2019transformer}
Z.~Dai, Z.~Yang, Y.~Yang, J.~Carbonell, Q.~V. Le, and R.~Salakhutdinov.
\newblock Transformer-xl: Attentive language models beyond a fixed-length context.
\newblock \emph{arXiv preprint arXiv:1901.02860}, 2019.

\bibitem[Dalla-Torre et~al.(2024)Dalla-Torre, Gonzalez, Mendoza-Revilla, Lopez~Carranza, Grzywaczewski, Oteri, Dallago, Trop, de~Almeida, Sirelkhatim, et~al.]{dalla2024nucleotide}
H.~Dalla-Torre, L.~Gonzalez, J.~Mendoza-Revilla, N.~Lopez~Carranza, A.~H. Grzywaczewski, F.~Oteri, C.~Dallago, E.~Trop, B.~P. de~Almeida, H.~Sirelkhatim, et~al.
\newblock Nucleotide transformer: building and evaluating robust foundation models for human genomics.
\newblock \emph{Nature Methods}, pages 1--11, 2024.

\bibitem[Devlin et~al.(2019)Devlin, Chang, Lee, and Toutanova]{devlin2019bert}
J.~Devlin, M.-W. Chang, K.~Lee, and K.~Toutanova.
\newblock Bert: Pre-training of deep bidirectional transformers for language understanding.
\newblock In \emph{Proceedings of the 2019 conference of the North American chapter of the association for computational linguistics: human language technologies, volume 1 (long and short papers)}, pages 4171--4186, 2019.

\bibitem[He et~al.(2020)He, Liu, Gao, and Chen]{he2020deberta}
P.~He, X.~Liu, J.~Gao, and W.~Chen.
\newblock Deberta: Decoding-enhanced bert with disentangled attention.
\newblock \emph{arXiv preprint arXiv:2006.03654}, 2020.

\bibitem[Ji(1999)]{ji_linguistics_1999}
S.~Ji.
\newblock The linguistics of {DNA}: words, sentences, grammar, phonetics, and semantics.
\newblock 870:\penalty0 411--417, 1999.
\newblock ISSN 0077-8923.

\bibitem[Ji et~al.(2021)Ji, Zhou, Liu, and Davuluri]{ji2021dnabert}
Y.~Ji, Z.~Zhou, H.~Liu, and R.~V. Davuluri.
\newblock {DNABERT: pre-trained Bidirectional Encoder Representations from Transformers model for DNA-language in genome}.
\newblock \emph{Bioinformatics}, 37\penalty0 (15):\penalty0 2112--2120, 2021.

\bibitem[Jolma et~al.(2015)Jolma, Yin, Nitta, Dave, Popov, Taipale, Enge, Kivioja, Morgunova, and Taipale]{jolma2015dna}
A.~Jolma, Y.~Yin, K.~R. Nitta, K.~Dave, A.~Popov, M.~Taipale, M.~Enge, T.~Kivioja, E.~Morgunova, and J.~Taipale.
\newblock Dna-dependent formation of transcription factor pairs alters their binding specificity.
\newblock \emph{Nature}, 527\penalty0 (7578):\penalty0 384--388, 2015.

\bibitem[Kelley et~al.(2016)Kelley, Snoek, and Rinn]{kelley2016basset}
D.~R. Kelley, J.~Snoek, and J.~L. Rinn.
\newblock Basset: learning the regulatory code of the accessible genome with deep convolutional neural networks.
\newblock \emph{Genome research}, 26\penalty0 (7):\penalty0 990--999, 2016.

\bibitem[Kitaev et~al.(2020)Kitaev, Kaiser, and Levskaya]{kitaev2020reformer}
N.~Kitaev, {\L}.~Kaiser, and A.~Levskaya.
\newblock Reformer: The efficient transformer.
\newblock \emph{arXiv preprint arXiv:2001.04451}, 2020.

\bibitem[Li et~al.(2023)Li, Gao, Zhou, Han, Xu, and Gao]{li_applications_2023}
Z.~Li, E.~Gao, J.~Zhou, W.~Han, X.~Xu, and X.~Gao.
\newblock Applications of deep learning in understanding gene regulation.
\newblock 3\penalty0 (1):\penalty0 100384, 2023.
\newblock ISSN 2667-2375.

\bibitem[Liang et~al.(2020)Liang, Bible, Liu, Zou, and Wei]{liang2020deepmicrobes}
Q.~Liang, P.~W. Bible, Y.~Liu, B.~Zou, and L.~Wei.
\newblock Deepmicrobes: taxonomic classification for metagenomics with deep learning.
\newblock \emph{NAR Genomics and Bioinformatics}, 2\penalty0 (1):\penalty0 lqaa009, 2020.

\bibitem[Liu et~al.(2024)Liu, Zhou, Chen, Liu, Huo, and Han]{liu2024exploring}
H.~Liu, S.~Zhou, P.~Chen, J.~Liu, K.-G. Huo, and L.~Han.
\newblock Exploring genomic large language models: Bridging the gap between natural language and gene sequences.
\newblock \emph{bioRxiv}, pages 2024--02, 2024.

\bibitem[Liutkus et~al.(2021)Liutkus, C{\i}fka, Wu, Simsekli, Yang, and Richard]{liutkus2021relative}
A.~Liutkus, O.~C{\i}fka, S.-L. Wu, U.~Simsekli, Y.-H. Yang, and G.~Richard.
\newblock Relative positional encoding for transformers with linear complexity.
\newblock In \emph{International Conference on Machine Learning}, pages 7067--7079. PMLR, 2021.

\bibitem[Loshchilov and Hutter(2017)]{loshchilov2017decoupled}
I.~Loshchilov and F.~Hutter.
\newblock Decoupled weight decay regularization.
\newblock \emph{arXiv preprint arXiv:1711.05101}, 2017.

\bibitem[Mao et~al.(2025)Mao, Tian, and Song]{mao2025dnazen}
L.~Mao, Y.~Tian, and Y.~Song.
\newblock Dnazen: Enhanced gene sequence representations via mixed granularities of coding units.
\newblock \emph{arXiv preprint arXiv:2505.02206}, 2025.

\bibitem[Mock et~al.(2022)Mock, Kretschmer, Kriese, B{\"o}cker, and Marz]{mock2022taxonomic}
F.~Mock, F.~Kretschmer, A.~Kriese, S.~B{\"o}cker, and M.~Marz.
\newblock Taxonomic classification of dna sequences beyond sequence similarity using deep neural networks.
\newblock \emph{Proceedings of the National Academy of Sciences}, 119\penalty0 (35):\penalty0 e2122636119, 2022.

\bibitem[Nguyen et~al.(2023)Nguyen, Poli, Faizi, Thomas, Wornow, Birch-Sykes, Massaroli, Patel, Rabideau, Bengio, et~al.]{nguyen2023hyenadna}
E.~Nguyen, M.~Poli, M.~Faizi, A.~Thomas, M.~Wornow, C.~Birch-Sykes, S.~Massaroli, A.~Patel, C.~Rabideau, Y.~Bengio, et~al.
\newblock Hyenadna: Long-range genomic sequence modeling at single nucleotide resolution.
\newblock \emph{Advances in neural information processing systems}, 36:\penalty0 43177--43201, 2023.

\bibitem[Nguyen et~al.(2024)Nguyen, Poli, Durrant, Kang, Katrekar, Li, Bartie, Thomas, King, Brixi, et~al.]{nguyen2024sequence}
E.~Nguyen, M.~Poli, M.~G. Durrant, B.~Kang, D.~Katrekar, D.~B. Li, L.~J. Bartie, A.~W. Thomas, S.~H. King, G.~Brixi, et~al.
\newblock Sequence modeling and design from molecular to genome scale with evo.
\newblock \emph{Science}, 386\penalty0 (6723):\penalty0 eado9336, 2024.

\bibitem[Ouyang et~al.(2022)Ouyang, Wu, Jiang, Almeida, Wainwright, Mishkin, Zhang, Agarwal, Slama, Ray, et~al.]{ouyang2022training}
L.~Ouyang, J.~Wu, X.~Jiang, D.~Almeida, C.~Wainwright, P.~Mishkin, C.~Zhang, S.~Agarwal, K.~Slama, A.~Ray, et~al.
\newblock Training {L}anguage {M}odels to {F}ollow {I}nstructions with {H}uman {F}eedback.
\newblock \emph{Advances in Neural Information Processing Systems}, 35:\penalty0 27730--27744, 2022.

\bibitem[Press et~al.(2021)Press, Smith, and Lewis]{press2021train}
O.~Press, N.~A. Smith, and M.~Lewis.
\newblock Train short, test long: Attention with linear biases enables input length extrapolation.
\newblock \emph{arXiv preprint arXiv:2108.12409}, 2021.

\bibitem[Qin et~al.(2022)Qin, Tian, Xia, and Song]{qin-2022-complementary}
H.~Qin, Y.~Tian, F.~Xia, and Y.~Song.
\newblock Complementary {L}earning of {A}spect {T}erms for {A}spect-based {S}entiment {A}nalysis.
\newblock In \emph{Proceedings of the 13th Language Resources and Evaluation Conference}, 2022.

\bibitem[Raffel et~al.(2020)Raffel, Shazeer, Roberts, Lee, Narang, Matena, Zhou, Li, and Liu]{raffel2020exploring}
C.~Raffel, N.~Shazeer, A.~Roberts, K.~Lee, S.~Narang, M.~Matena, Y.~Zhou, W.~Li, and P.~J. Liu.
\newblock Exploring the limits of transfer learning with a unified text-to-text transformer.
\newblock \emph{Journal of machine learning research}, 21\penalty0 (140):\penalty0 1--67, 2020.

\bibitem[Rozowsky et~al.(2023)Rozowsky, Gao, Borsari, Yang, Galeev, Gürsoy, Epstein, Xiong, Xu, Li, Liu, Yu, Berthel, Chen, Navarro, Sun, Wright, Chang, Cameron, Shoresh, Gaskell, Drenkow, Adrian, Aganezov, Aguet, Balderrama-Gutierrez, Banskota, Corona, Chee, Chhetri, Martins, Danyko, Davis, Farid, Farrell, Gabdank, Gofin, Gorkin, Gu, Hecht, Hitz, Issner, Jiang, Kirsche, Kong, Lam, Li, Li, Li, Lin, Luo, Mackiewicz, Meng, Moore, Mudge, Nelson, Nusbaum, Popov, Pratt, Qiu, Ramakrishnan, Raymond, Salichos, Scavelli, Schreiber, Sedlazeck, See, Sherman, Shi, Shi, Sloan, Strattan, Tan, Tanaka, Vlasova, Wang, Werner, Williams, Xu, Yan, Yu, Zaleski, Zhang, Ardlie, Cherry, Mendenhall, Noble, Weng, Levine, Dobin, Wold, Mortazavi, Ren, Gillis, Myers, Snyder, Choudhary, Milosavljevic, Schatz, Bernstein, Guigó, Gingeras, and Gerstein]{rozowsky_en-tex_2023}
J.~Rozowsky, J.~Gao, B.~Borsari, Y.~T. Yang, T.~Galeev, G.~Gürsoy, C.~B. Epstein, K.~Xiong, J.~Xu, T.~Li, J.~Liu, K.~Yu, A.~Berthel, Z.~Chen, F.~Navarro, M.~S. Sun, J.~Wright, J.~Chang, C.~J.~F. Cameron, N.~Shoresh, E.~Gaskell, J.~Drenkow, J.~Adrian, S.~Aganezov, F.~Aguet, G.~Balderrama-Gutierrez, S.~Banskota, G.~B. Corona, S.~Chee, S.~B. Chhetri, G.~C.~C. Martins, C.~Danyko, C.~A. Davis, D.~Farid, N.~P. Farrell, I.~Gabdank, Y.~Gofin, D.~U. Gorkin, M.~Gu, V.~Hecht, B.~C. Hitz, R.~Issner, Y.~Jiang, M.~Kirsche, X.~Kong, B.~R. Lam, S.~Li, B.~Li, X.~Li, K.~Z. Lin, R.~Luo, M.~Mackiewicz, R.~Meng, J.~E. Moore, J.~Mudge, N.~Nelson, C.~Nusbaum, I.~Popov, H.~E. Pratt, Y.~Qiu, S.~Ramakrishnan, J.~Raymond, L.~Salichos, A.~Scavelli, J.~M. Schreiber, F.~J. Sedlazeck, L.~H. See, R.~M. Sherman, X.~Shi, M.~Shi, C.~A. Sloan, J.~S. Strattan, Z.~Tan, F.~Y. Tanaka, A.~Vlasova, J.~Wang, J.~Werner, B.~Williams, M.~Xu, C.~Yan, L.~Yu, C.~Zaleski, J.~Zhang, K.~Ardlie, J.~M. Cherry, E.~M. Mendenhall, W.~S. Noble, Z.~Weng, M.~E.
  Levine, A.~Dobin, B.~Wold, A.~Mortazavi, B.~Ren, J.~Gillis, R.~M. Myers, M.~P. Snyder, J.~Choudhary, A.~Milosavljevic, M.~C. Schatz, B.~E. Bernstein, R.~Guigó, T.~R. Gingeras, and M.~Gerstein.
\newblock The {EN}-{TEx} resource of multi-tissue personal epigenomes \& variant-impact models.
\newblock 186\penalty0 (7):\penalty0 1493--1511.e40, 2023.
\newblock ISSN 0092-8674, 1097-4172.
\newblock Publisher: Elsevier.

\bibitem[Sanabria et~al.(2024)Sanabria, Hirsch, Joubert, and Poetsch]{sanabria2024dna}
M.~Sanabria, J.~Hirsch, P.~M. Joubert, and A.~R. Poetsch.
\newblock Dna language model grover learns sequence context in the human genome.
\newblock \emph{Nature Machine Intelligence}, 6\penalty0 (8):\penalty0 911--923, 2024.

\bibitem[Schiff et~al.(2024)Schiff, Kao, Gokaslan, Dao, Gu, and Kuleshov]{schiff2024caduceus}
Y.~Schiff, C.-H. Kao, A.~Gokaslan, T.~Dao, A.~Gu, and V.~Kuleshov.
\newblock Caduceus: Bi-directional equivariant long-range dna sequence modeling.
\newblock \emph{arXiv preprint arXiv:2403.03234}, 2024.

\bibitem[Searls(2002)]{searls_language_2002}
D.~B. Searls.
\newblock The language of genes.
\newblock 420\penalty0 (6912):\penalty0 211--217, 2002.
\newblock ISSN 1476-4687.
\newblock Publisher: Nature Publishing Group.

\bibitem[Sennrich et~al.(2015)Sennrich, Haddow, and Birch]{sennrich2015neural}
R.~Sennrich, B.~Haddow, and A.~Birch.
\newblock Neural machine translation of rare words with subword units.
\newblock \emph{arXiv preprint arXiv:1508.07909}, 2015.

\bibitem[Sharon et~al.(2012)Sharon, Kalma, Sharp, Raveh-Sadka, Levo, Zeevi, Keren, Yakhini, Weinberger, and Segal]{sharon2012inferring}
E.~Sharon, Y.~Kalma, A.~Sharp, T.~Raveh-Sadka, M.~Levo, D.~Zeevi, L.~Keren, Z.~Yakhini, A.~Weinberger, and E.~Segal.
\newblock Inferring gene regulatory logic from high-throughput measurements of thousands of systematically designed promoters.
\newblock \emph{Nature biotechnology}, 30\penalty0 (6):\penalty0 521--530, 2012.

\bibitem[Su et~al.(2024)Su, Ahmed, Lu, Pan, Bo, and Liu]{su2024roformer}
J.~Su, M.~Ahmed, Y.~Lu, S.~Pan, W.~Bo, and Y.~Liu.
\newblock Roformer: Enhanced transformer with rotary position embedding.
\newblock \emph{Neurocomputing}, 568:\penalty0 127063, 2024.

\bibitem[Tian et~al.(2021)Tian, Chen, Qin, and Song]{tian-etal-2021-federated}
Y.~Tian, G.~Chen, H.~Qin, and Y.~Song.
\newblock Federated {C}hinese {W}ord {S}egmentation with {G}lobal {C}haracter {A}ssociations.
\newblock In \emph{Findings of the Association for Computational Linguistics: ACL-IJCNLP 2021}, 2021.

\bibitem[Touvron et~al.(2023)Touvron, Lavril, Izacard, Martinet, Lachaux, Lacroix, Rozi{\`e}re, Goyal, Hambro, Azhar, et~al.]{touvron2023llama}
H.~Touvron, T.~Lavril, G.~Izacard, X.~Martinet, M.-A. Lachaux, T.~Lacroix, B.~Rozi{\`e}re, N.~Goyal, E.~Hambro, F.~Azhar, et~al.
\newblock {LLaMA: Open and Efficient Foundation Language Models}.
\newblock \emph{arXiv preprint arXiv:2302.13971}, 2023.

\bibitem[Vaswani et~al.(2017)Vaswani, Shazeer, Parmar, Uszkoreit, Jones, Gomez, Kaiser, and Polosukhin]{vaswani2017attention}
A.~Vaswani, N.~Shazeer, N.~Parmar, J.~Uszkoreit, L.~Jones, A.~N. Gomez, {\L}.~Kaiser, and I.~Polosukhin.
\newblock Attention is {A}ll {Y}ou {N}eed.
\newblock \emph{Advances in neural information processing systems}, 30, 2017.

\bibitem[Wen et~al.(2014)Wen, Chan, Yau, He, and Yau]{wen2014k}
J.~Wen, R.~H. Chan, S.-C. Yau, R.~L. He, and S.~S. Yau.
\newblock K-mer natural vector and its application to the phylogenetic analysis of genetic sequences.
\newblock \emph{Gene}, 546\penalty0 (1):\penalty0 25--34, 2014.

\bibitem[Wu et~al.(2025)Wu, Li, Li, Fu, Feng, Ye, Xiong, and Wang]{wu2025generator}
W.~Wu, Q.~Li, M.~Li, K.~Fu, F.~Feng, J.~Ye, H.~Xiong, and Z.~Wang.
\newblock Generator: A long-context generative genomic foundation model.
\newblock \emph{arXiv preprint arXiv:2502.07272}, 2025.

\bibitem[Yang et~al.(2025)Yang, Li, Yang, Zhang, Hui, Zheng, Yu, Gao, Huang, Lv, et~al.]{yang2025qwen3}
A.~Yang, A.~Li, B.~Yang, B.~Zhang, B.~Hui, B.~Zheng, B.~Yu, C.~Gao, C.~Huang, C.~Lv, et~al.
\newblock Qwen3 technical report.
\newblock \emph{arXiv preprint arXiv:2505.09388}, 2025.

\bibitem[Yoon et~al.(2002)Yoon, Liyanarachchi, Wright, Davuluri, Lockman, de~la Chapelle, and Pellegata]{yoon_gene_2002}
H.~Yoon, S.~Liyanarachchi, F.~A. Wright, R.~Davuluri, J.~C. Lockman, A.~de~la Chapelle, and N.~S. Pellegata.
\newblock Gene expression profiling of isogenic cells with different {TP}53 gene dosage reveals numerous genes that are affected by {TP}53 dosage and identifies {CSPG}2 as a direct target of p53.
\newblock 99\penalty0 (24):\penalty0 15632--15637, 2002.
\newblock Publisher: Proceedings of the National Academy of Sciences.

\bibitem[Zhang et~al.(2023)Zhang, Zhang, and Zhao]{zhang2023dnagpt}
D.~Zhang, W.~Zhang, and U.~Zhao.
\newblock {DNAGPT: A Generalized Pre-trained Tool for DNA Sequence Analysis}.
\newblock \emph{arXiv preprint arXiv:2307.05628}, 2023.

\bibitem[Zhou and Troyanskaya(2015)]{zhou2015predicting}
J.~Zhou and O.~G. Troyanskaya.
\newblock Predicting effects of noncoding variants with deep learning--based sequence model.
\newblock \emph{Nature methods}, 12\penalty0 (10):\penalty0 931--934, 2015.

\bibitem[Zhou et~al.(2023)Zhou, Ji, Li, Dutta, Davuluri, and Liu]{zhou2023dnabert}
Z.~Zhou, Y.~Ji, W.~Li, P.~Dutta, R.~Davuluri, and H.~Liu.
\newblock {DNABERT-2: Efficient foundation model and benchmark for multi-species genome}.
\newblock \emph{arXiv preprint arXiv:2306.15006}, 2023.

\end{thebibliography}

\end{document}